\documentclass{article}

\usepackage[in]{fullpage}  
\usepackage{mathpazo} 

\usepackage{microtype}
\usepackage{graphicx}
\usepackage{subfigure}
\usepackage{booktabs} 
\usepackage{hyperref}

\usepackage{amsmath}
\usepackage{amssymb}
\usepackage{mathtools}
\usepackage{amsthm}

\usepackage{enumitem}
\usepackage{tcolorbox}
\usepackage[capitalize,noabbrev]{cleveref}

\AtBeginDocument{\setlength\abovedisplayskip{4pt}}
\AtBeginDocument{\setlength\belowdisplayskip{4pt}}

\definecolor{red}{rgb}{1, 0, 0}

\newcommand{\citep}{\cite}

\theoremstyle{plain}

\theoremstyle{definition}

\theoremstyle{remark}

\usepackage{algorithm,algorithmicx,algpseudocode}
\usepackage[utf8]{inputenc} 
\usepackage[T1]{fontenc}    
\usepackage{hyperref}       
\usepackage{url}            
\usepackage{booktabs}       
\usepackage{amsfonts}       
\usepackage{nicefrac}       
\usepackage{microtype}      
\usepackage{xcolor}         
\usepackage{wrapfig}

\title{BWArea Model: Learning World Model, Inverse Dynamics, and Policy for Controllable Language Generation}

\author{%
  Chengxing Jia$^{1,2,5}$, Pengyuan Wang$^{1,2,5}$, Ziniu Li$^{3,4}$, Yi-Chen Li$^{1,2,5}$, Zhilong Zhang$^{1,2,5}$,\\
  Nan Tang$^{1,2,5}$, Yang Yu$^{1,2,5}$\thanks{Corresponding author: yuy@nju.edu.cn}\\
  \small $^1$National Key Laboratory for Novel Software Technology, Nanjing University\\
  \small $^2$School of Artificial Intelligence, Nanjing University\\
  \small $^3$School of Data Science, The Chinese University of Hong Kong, Shenzhen\\
\small $^4$Shenzhen Research Institute of Big Data\\
\small $^5$Polixir.ai \\
}
\date{}

\begin{document}

\maketitle

\begin{abstract}

Large language models (LLMs) have catalyzed a paradigm shift in natural language processing, yet their limited controllability poses a significant challenge for downstream applications. We aim to address this by drawing inspiration from the neural mechanisms of the human brain, specifically Broca's and Wernicke's areas, which are crucial for language generation and comprehension, respectively.
In particular, Broca's area receives cognitive decision signals from Wernicke's area, treating the language generation as an intricate decision-making process, which differs from the fully auto-regressive language generation of existing LLMs. In a similar vein, our proposed system, the \emph{BWArea model}, conceptualizes language generation as a decision-making task. This model has three components: a language world model, an inverse dynamics model, and a cognitive policy. Like Wernicke's area, the inverse dynamics model is designed to deduce the underlying cognitive intentions, or latent actions, behind each token. The BWArea model is amenable to both pre-training and fine-tuning like existing LLMs. With 30B clean pre-training tokens, we have trained a BWArea model, which achieves competitive performance with LLMs of equal size (1B parameters). Unlike fully auto-regressive LLMs, its pre-training performance does not degenerate if dirty data unintentionally appears. This shows the advantage of a decomposed structure of BWArea model in reducing efforts in laborious data selection and labeling. Finally, we reveal that the BWArea model offers enhanced controllability via fine-tuning the cognitive policy with downstream reward metrics, thereby facilitating alignment with greater simplicity. On 9 out of 10 tasks from two suites, TextWorld and BigBench Hard, our method shows superior performance to auto-regressive LLMs.

\end{abstract}

\vspace{-0.3cm}
\section{Introduction}
\vspace{-0.2cm}

Large language models (LLMs) have become the cornerstone of modern natural language processing, providing the foundation for powerful approaches to understanding and generating human language. They have revolutionized numerous applications such as machine translation and question-answering \citep{devlin2018bert,brown2020language}. However, despite their widespread success and utility, the issue of controllability is one of the most pressing challenges in this field~\citep{guo2024cold, bhargava2023s}. Current LLMs generate language in an auto-regressive manner, which assumes the next token depends solely on the previous tokens, thus the generation process is hard to interfere with. Even if alignment and prompt optimization methods were developed, they still lack the nuanced control that is desirable for many applications, such as those requiring precise alignment with human values or adherence to specific content guidelines \citep{bender2021dangers}.



\begin{figure}
\vspace{-8mm}
    \centering
    \includegraphics[width=0.99\textwidth]{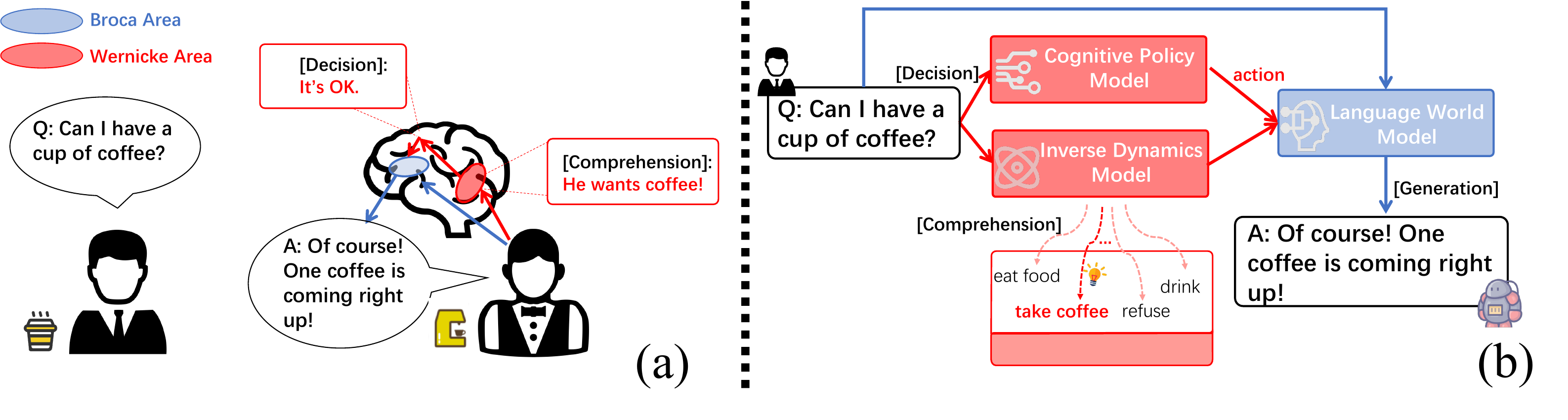}
    \vskip -0.1in
    \caption{An example of how our BWArea model mimics the human brain for language processing.}
    \label{fig:bw}
    \vspace{-4mm}
\end{figure}

In this paper, we aim to tackle the challenge of controllability. Our approach draws inspiration from the brain's language processing centers: the Broca's and Wernicke's areas. These regions are integral to language generation and comprehension \citep{amunts2004broca}. Broca's area is responsible for language generation. When we write, it helps form words and construct sentences. Unlike the auto-regressive nature of existing LLMs, Broca's area engages in language generation under the governance of cognitive decisions~(e.g. decisions from the frontal lobe~\citep{collins2012reasoning}
), suggesting that language generation is deeply rooted in decision-making \citep{friederici2011brain}. 
On the other hand, Wernicke's area is responsible for language comprehension. When we read text, Wernicke's area helps us comprehend the meaning. Their functionality offers a blueprint for a more controllable language generation system and has profound implications for the design of our proposed model, the BWArea model; see Figure~\ref{fig:bw}.



Our BWArea model reimagines language generation as a decision-making task. It is an integration of three interrelated components: a language world model, an inverse dynamics model, and a cognitive policy.
The language world model generates tokens driven by cognitive decisions, the inverse dynamics model infers the latent cognitive decisions that propel each token's generation.
Using the inverse dynamics model, we transform language sequences into decision trajectories by filling in the latent cognitive decisions. This conversion enables the application of behavior cloning to train the language world model and cognitive policy. By treating language as a series of decisions, we facilitate a learning process that mirrors the acquisition of language in the human brain.


The BWArea model can undergo both pre-training and fine-tuning, similar to existing LLMs. We have trained a BWArea model from scratch. The world model and policy model have 1B parameters, while the inverse model is lightweight with 0.5B parameters. An auto-regressive LLM with equal size (1B parameters) was also trained for fair comparison. With 30B \emph{clean} pre-training tokens from the Slimpajama and StarCoder datasets, our model shows competitive evaluation performance with the auto-regressive LLM on common benchmarks such as MMLU, DROP, BBH, and TruthfulQA. Nevertheless, their performance differs when dealing with \emph{dirty} data where labels are noisy. In particular, with an additional 1B dirty tokens, the evaluation performance of our method does not degenerate and even improves by 3\%. In contrast, the performance of the auto-regressive LLM is reduced by 1.3\%. This shows that our method, due to its decomposed structure, has advantages in reducing the efforts of laborious data cleaning and labeling.

Furthermore, we also verify that the BWArea model has superior controllability in downstream applications. Technically, the cognitive policy selects a latent action while the world model predicts tokens conditioned on it, thus reducing the predictive variance for token generation. This allows for flexible adjustment of the cognitive policy via reward maximization using reinforcement learning methods. Empirically, we have found that the BWArea Model outperforms existing LLMs on several types of RL tasks, including three tasks from the TextWorld suite~\citep{cote18textworld}, and 6 out of 7 tasks from the BigBench Hard suite~\citep{suzgun2022challenging}. Collectively, these results show promising performance in treating language processing as a decision-making task, and we hope this can shed light on the development of controllable language generation systems.

\vspace{-0.1cm}
\section{Related Work}
\vspace{-0.1cm}

\vspace{-0.1cm}
\subsection{Large Language Models}
\vspace{-0.1cm}

The architecture of current Large Language Models (LLMs) has seen significant evolution since the introduction of the Transformer model \cite{vaswani2017attention}. The Transformer's self-attention mechanism allows for the parallel processing of sequences, which is a key factor contributing to the scaling capabilities of LLMs. One of the most influential architectures in this domain is BERT~\cite{devlin2018bert}, which utilizes a bidirectional Transformer~\citep{yang2019xlnet, lan2019albert} to pre-train deep bidirectional representations by jointly conditioning on both left and right context in all layers.
Following BERT, several models like GPT~\citep{radford2018improving}, and GPT-2~\citep{radford2019language} extended the use of Transformers in language modeling tasks by focusing on the generative pre-training and fine-tuning paradigm. Brown et al.~\cite{brown2020language} further scaled this approach with GPT-3, a model with 175 billion parameters, demonstrating that larger models can exhibit few-shot learning capabilities, where the model can perform tasks given very few examples. Recently, several other large language models have also embraced such a learning structure~\citep{tang23glm, colin20t5, meta23llama}.   

However, these methods have primarily focused on predicting the next token in a corpus, which can limit their ability to decide what to generate and may lead to what is commonly known as model illusion~\citep{and24ill}. To improve the controllability and efficiency of LLMs, prompt-based methods~\citep{liu2023pre, han2022prompt, wei2022chain, yao2024tree} have emerged as a pivotal technique. This method leverages the pre-trained knowledge encoded in the LLMs and has been shown to be effective even with few-shot examples. Prompts can be manually designed or automatically generated, and recent works have explored methods for optimizing prompts to better guide model behavior \cite{liu2021pre}. \citep{edward23llm} also involves a flow-based~\citep{bengio23gflow} method to model the prompt for the language generation. Alongside architectural advancements, there is a growing focus on aligning LLMs with human values. Reinforcement Learning from Human Feedback~(RLHF) has been proposed as a method to fine-tune models based on human preferences~\cite{christiano2017deep}. Another approach is Deep Preference Optimization (DPO), which iteratively refines model outputs to align with human values and preferences \cite{hendrycks2021aligning}. These methods aim to ensure that LLMs act in ways that are beneficial and non-harmful, according to human judgment.

\vspace{-0.1cm}
\subsection{World Models}
\vspace{-0.1cm}

World models in reinforcement learning (RL)~\citep{sutton2018reinforcement}, also called environment models and action models, are neural networks that simulate the environment with which an agent interacts. These models are then used to simulate trajectories, enabling the agent to plan and learn policies without interacting with the actual environment, thus saving valuable samples~\citep{janner2019trust, chua2018deep, luo2018algorithmic}.

\citep{ha2018world}~introduced the concept of world models by combining variational autoencoders~(VAEs)~\citep{kingma2013auto} for learning a compressed spatial representation, with recurrent neural networks (RNNs) for temporal predictions.
The use of world models has been extended by~\citep{hafner2019dream} with the Dreamer algorithm, which learns a latent dynamics model that can efficiently predict ahead for multiple time-steps in a compact latent space. Dreamer-V2~\citep{hafner2020dreamerv2} further improved the sample efficiency and scalability of this approach.
The MuZero algorithm~\citep{schrittwieser2020mastering}, which combines tree-based search with a learned dynamics model~\citep{oh2017value, anthony2017thinking}, has achieved state-of-the-art performance in various domains by learning a model that directly predicts the value, policy, and reward from the current state and action. In real-world scenarios where the collection of actions is lacking, there are additional works~\citep{gao23inv, SeoLJA22af} that extract latent actions from the dataset to control the state.

\vspace{-0.1cm}
\section{BWArea Model}
\vspace{-0.1cm}




\vspace{-0.1cm}
\subsection{Model Design}
\vspace{-0.1cm}

We are motivated by the working mechanism of the Broca and Wernicke areas in the human brain. We build an inverse dynamics model and a world model that mimic the language understanding process of the Wernicke area (consider when we are listening to or reading some text). The inverse dynamics model infers a latent signal, referred to as action in the terminology of RL, that helps predict the next token generation. The main feature is that this signal has certain semantic meanings, which eases token generation (i.e., by reducing the predictive covariance). Later on, we introduce a cognitive policy that operates with the latent action found by the inverse dynamics model to mimic the language generation process of the Broca area. At the inference stage (consider when we are about to speak), the cognitive policy selects a latent action and asks the world model to decode it into discrete tokens.

The above three models are implemented with the Transformer architecture \citep{vaswani2017attention} and go through the pre-training and fine-tuning stages similar to classical auto-regressive LLMs, as explained below.





\vspace{-0.1cm}
\subsubsection{Language World Model}
\vspace{-0.1cm}


The language world model $p_{\texttt{world}}$ aims to predict the next token based on the context and a latent action, with the latter being the main difference from auto-regressive LLMs. Let $x_{1:t}$ be a sequence of tokens $(x_1, \ldots, x_t)$ and $x_{t+1}$ be the next token in a sentence. In our formulation, the next token prediction is based on $x_{t+1} \sim p_{\texttt{world}}(x_{1:t}, a_{1:t})$; in RL terminology, it serves as the transition model. By this additional condition, the predictive variance of the next token can be reduced. We will discuss how to identify this latent action later, but for now, let us explain the details of this world model.

\begin{figure}[t]
\vspace{-12mm}
    \centering
    \includegraphics[width=0.96\textwidth]{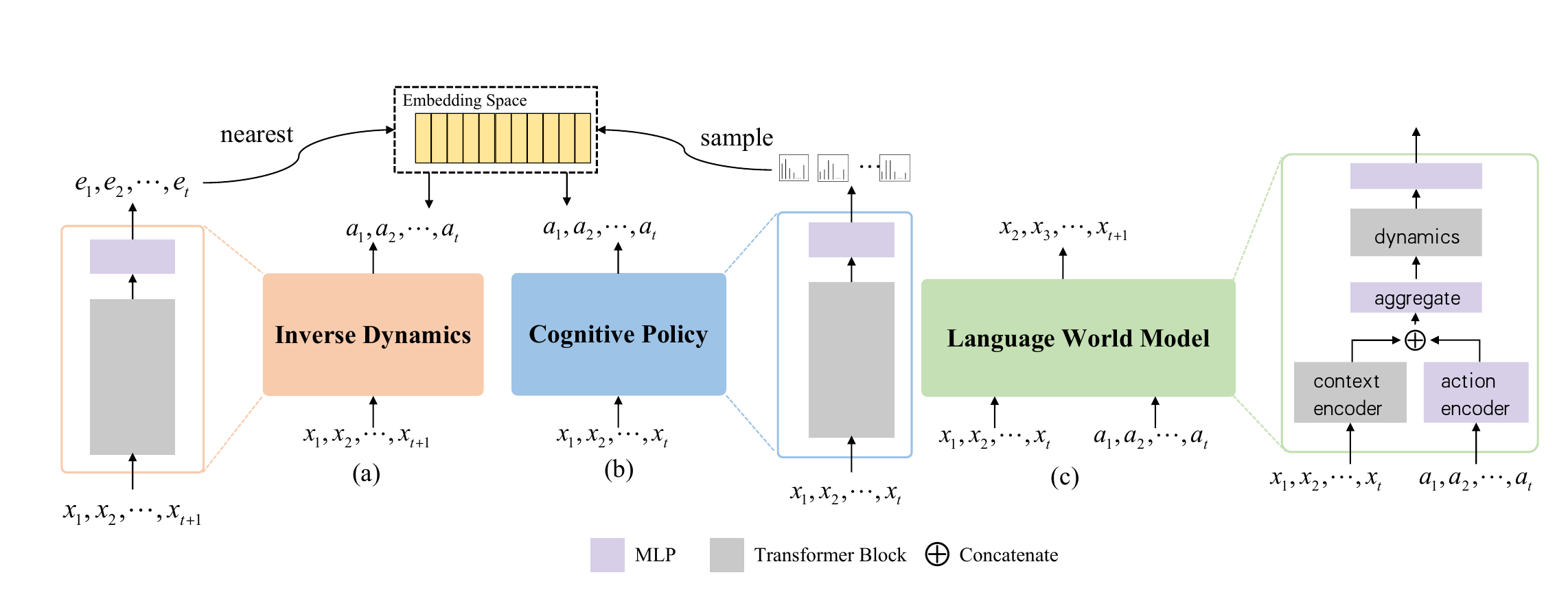}
    \vskip -0.1in
    \caption{Framework of our architecture. (a) Inverse Dynamics Model: input the context $(x_1, \ldots, x_t)$ with the future $x_{t+1}$ to output the latent action $a_t$. (b) Policy Model: input the context $(x_1, \ldots, x_t)$ without future to obtain the current action categorical distribution. (c) Language World Model: input the context $(x_1, \ldots, x_t)$ and latent action $(a_1, \ldots, a_t)$ to predict the next token.}
    \label{fig:fw}
    \vspace{-4mm}
\end{figure}

\vspace{-0.1cm}
\begin{itemize}[topsep=1pt,parsep=1pt,partopsep=1pt, leftmargin=*]
    \item \textbf{Context encoder} $f_{\texttt{world}}^{\texttt{context}}$, a transformer-based structure which inputs the context $x_{1:t}$ and outputs their embeddings at each timestep $e^s_{1:t}$;
    \item \textbf{Action encoder} $f_{\texttt{world}}^{\texttt{action}}$, a linear layer which inputs the action sequence $a_{1:t}$ and outputs their embeddings at each timestep $e^a_{1:t}$;
    \item \textbf{Aggregate module} $f_{\texttt{world}}^{\texttt{agg}}$, a linear layer which inputs the embeddings of context $e^s_t$ and action $e^a_t$ at timestep $t$ to input the joint embeddings $e_t$;
    \item \textbf{Dynamics module} $f_{\texttt{world}}^{\texttt{dyna}}$, a transformer-based structure which inputs the sequence of embeddings $e_{1:t}$ and outputs the next-token prediction, which is the categorical logits of token $x_{2:t+1}$.
\end{itemize}
\vspace{-0.1cm}
Therefore, the inference can be expressed by $f_{\texttt{world}}^{\texttt{dyna}} \circ f_{\texttt{world}}^{\texttt{agg}}([f_{\texttt{world}}^{\texttt{context}}(x_{1:t}), f_{\texttt{world}}^{\texttt{action}}(a_{1:t})])$, and we also give an illustration of the inference process in Figure~\ref{fig:fw}(c).

\vspace{-0.1cm}
\subsubsection{Inverse Dynamics Model}
\vspace{-0.1cm}


We introduce an inverse dynamics model, $f_{\texttt{inverse}}$, to bridge the language understanding and generation. For understanding, the model is trained to infer a latent action $a_t$ from the observed data. For generation, it constructs the action space from which the cognitive policy can select actions. We hypothesize that the latent action $a_t$ aids in predicting the subsequent tokens that can be retrospectively inferred (i.e., hindsight). In our work, we investigate a specific inverse dynamics model that leverages the next token $x_{t+1}$ as well as the contextual history $x_{1:t}$ to infer the latent action, i.e., $a_t = f_{\texttt{inverse}}(x_{1:t}, x_{t+1})$. Exploration of other scenarios, such as utilizing future $k$ tokens $x_{t+1:t+k}$ to determine the latent action, is left for future research.



We highlight that it is important to learn a moderate action space. The classical auto-regressive LLMs in our formulation correspond to the case where the action space is a singleton (with a dummy action), with these models predicting the next token solely on the context. On the other extreme, if the action space corresponds directly to the token space, selecting an action would be equivalent to selecting a token, resulting in an overwhelming decision-making burden. This is not our desired approach.

To address the above challenge, we implement a vector-quantized varational-auto-encoder (VQ-VAE) \citep{van2017neural}, which learns discrete feature representations from data. In our setting, this VQ-VAE encoder has the following modules:
\vspace{-0.1cm}
\begin{itemize}[topsep=1pt,parsep=1pt,partopsep=1pt, leftmargin=*]
    \item \textbf{Context encoder} $f_{\texttt{inverse}}^{\texttt{context}}$, a transformer-based structure which inputs the context $x_{1:t+1}$ and outputs their embeddings at each timestep $e_{1:t}$;
    \item \textbf{Action codebook} $f_{\texttt{inverse}}^{\texttt{code}}$, a finite codebook $\mathcal{C} = \{c^i\}_{i=1}^{N}$ with $N$ codes~\footnote{In our experiments, The token space size is 32k while the action space size $N$ is 64.}. A code $c^{i}$ corresponds to a latent action in our design;
\end{itemize}
\vspace{-0.1cm}

The VQ-VAE model takes the $x_{1:t+1}$ as inputs, and outputs an embedding $e_{t}$ by $f^{c}_{\texttt{policy}}(x_{1:t+1})$. Then mapped to an action by $a_t = f_{\texttt{inverse}}^{\texttt{code}}(e_t)$. In this way, we effectively restrict the action space to be compact. Note that this codebook is learnable. We also give an illustration in Figure~\ref{fig:fw}(a).

\vspace{-0.1cm}
\subsubsection{Cognitive Policy Model}
\vspace{-0.1cm}


In the above part, we illustrate the inverse dynamics model builds a latent action space and the world model predicts the next token $x_{t+1}$ given on the context $x_{1:t}$  and action $a_t$. Now, we implement a cognitive (decision) policy model $\pi_{\texttt{policy}}$ to select latent action $a_t$ for manipulating the token generation process.
Specifically, this policy model takes the context $x_{1:t}$ as its input and generates an action $a_t \in \mathcal{C}$ from a categorical distribution: $a_t \sim \pi_{\texttt{policy}}(x_{1:t}).$

We highlight that the goal of the policy model is to perform specific down-stream tasks via reward maximization. It is trained offline with policy gradient methods \citep{li2023remax} in our paper but it may be also combined with online optimization methods such as model predictive control~\citep{schwenzer2021review, hansen2022temporal}. Compared with auto-regressive LLMs that operate in the raw token space, the policy model introduced here is more flexible in controlling token generation by the world model.



        






\algrenewcommand\algorithmicindent{0.5em}%
\begin{figure}[t]
\begin{minipage}[t]{0.495\textwidth}
\begin{algorithm}[H]
  \caption{Pre-training} \label{alg:pre}
  \small
  \begin{algorithmic}[1]

     \Statex \textbf{Input:} Pretraining data $\mathcal{D}_{\texttt{pretrain}}$ and iters $K_{\texttt{pre}}$.

     \State \# Step 1
     \For{$t=1, \dotsc, K_{\texttt{pre}}$}
      \State Sample a batch of $x_{1:T}$ from $\mathcal{D}_{\texttt{pretrain}}$.
      \State Learn $\theta_{\texttt{world}}$ and $\theta_{\texttt{inverse}}$ by Equation~\ref{eq:pre1}.
    \EndFor

    \State \# Step 2
     \For{$t=1, \dotsc, K_{\texttt{pre}}$}
      \State Sample a batch of $x_{1:T}$ from $\mathcal{D}_{\texttt{pretrain}}$.
      \State Compute action target $a_{1:T}$ by $f_{\texttt{inverse}}$.
      \State Learn $\theta_{\texttt{policy}}$ by Equation~\ref{eq:pre2}.
    \EndFor
    
  \end{algorithmic}
\end{algorithm}
\end{minipage}
\hfill
\begin{minipage}[t]{0.495\textwidth}
\begin{algorithm}[H]
  \caption{Supervised Fine-Tuning} \label{alg:sft}
  \small
  \begin{algorithmic}[1]
    \Statex \textbf{Input:} SFT data $\mathcal{D}_{\texttt{sft}}$ and iterations $K_{\texttt{sft}}$.
    

     \State \# Step 1
     \For{$t=1, \dotsc, K_{\texttt{sft}}$}
      \State Sample a batch of $x_{1:T}$ from $\mathcal{D}_{\texttt{pretrain}}$.
      \State Learn $\theta_{\texttt{world}}$ and $\theta_{\texttt{inverse}}$ by Equation~\ref{eq:pre1} with $\mathcal{D}_{\texttt{sft}}$.
    \EndFor

    \State \# Step 2
     \For{$t=1, \dotsc, K_{\texttt{sft}}$}
      \State Sample a batch of $x_{1:T}$ from $\mathcal{D}_{\texttt{pretrain}}$.
      \State Compute action targets $a_{\textcolor{red}{p+1}:T}$ by $f_{\texttt{inverse}}$.
      \State Learn $\theta_{\texttt{policy}}$ by Equation~\ref{eq:pre2} with $\mathcal{D}_{\texttt{sft}}$.
    \EndFor
  \end{algorithmic}
\end{algorithm}
\end{minipage}
\vspace{-1em}
\end{figure}

\vspace{-0.15cm}
\subsection{Model Inference}
\vspace{-0.15cm}

We now introduce the inference process. For a prompt $x_{1:t}$, the next-token $x_{t+1}$ is generated by:
\begin{equation}
\label{eq:gen}
    \begin{split}
        x_{t+1} &\sim p_{\texttt{world}}(x_{t+1} | x_{1:t}, a_{1:t}; \theta_{\texttt{world}}), \\
    a_{i} &= f_{\texttt{inverse}}(x_{1:i+1}; \theta_{\texttt{inverse}}), \quad \forall i = 0, \ldots t-1 \\
    a_t &\sim \pi_{\texttt{policy}}(x_{1:t}; \theta_{\texttt{policy}}). 
    \end{split}
\end{equation}
That is, we first employ the inverse dynamics model to infer the action $a_{1:t-1}$ based on $x_{1:t}$, representing a comprehension of the language within the sentence. Subsequently, we execute action $a_t$ based on the policy, representing a specific intention. The contexts $x_{1:t}$ and actions $a_{1:t}$ are then input into the world language model to generate the next token $x_{t+1}$.

\vspace{-0.15cm}
\subsection{Model Training}
\vspace{-0.15cm}


The designed models go through three training stages. First, we pre-train the models to acquire basic language proficiency. Second, we engage in supervised fine-tuning (SFT) to improve the model's ability to follow instructions. Third, we utilize a reinforcement learning (RL) method to adapt the policy to finish specific tasks. For ease presetation, we use the symbols $\theta_{\texttt{world}}$, $\theta_{\texttt{inverse}}$, and $\theta_{\texttt{policy}}$ to denote the training parameters of the world model, inverse dynamics, and policy model, respectively.

\vspace{-0.1cm}
\subsubsection{Pre-training}
\vspace{-0.1cm}

In the pre-training stage, we are given a large corpus of texts, which provides rich knowledge for the acquisition of language decision-making. The pre-training dataset $\mathcal{D}_{\texttt{pretrain}}$ consists of numerous document segments. Each segment comprises a sequence of tokens as $x_{1:T}$. We extensively train the world model, inverse dynamics model, and policy model. The pre-training has two steps:

\textbf{Step 1}: we jointly train the inverse dynamics model and world model: 
\begin{equation}
\label{eq:pre1}
    \begin{split}
        \min_{\theta_{\texttt{world}}, \theta_{\texttt{inverse}}} \mathcal{L}_{\texttt{predict}}(\mathcal{D}_{\texttt{pretrain}}; \theta_{\texttt{world}}, \theta_{\texttt{inverse}}) + \beta \mathcal{L}_{\texttt{VQ}}(\mathcal{D}_{\texttt{pretrain}}; \theta_{\texttt{inverse}}) 
    \end{split}
\end{equation}
The objective $\mathcal{L}_{\texttt{predict}}$ is to predict the next token: 
\begin{align*}
 \mathcal{L}_{\texttt{predict}} = -\mathbb{E}_{x_{1:T} \sim \mathcal{D}_{\texttt{pretrain}}} \left[ \sum_{t=1}^{T} \log p_{\texttt{world}}(x_{t+1} | x_{1:t}, a_{1:t}, \theta_{\texttt{world}})] \right].  
\end{align*}
where $a_t = f_{\texttt{inverse}}(x_{1:t}, x_{t+1}; \theta_{\texttt{inverse}})$. The term $\mathcal{L}_{\texttt{VQ}}$ is the standard VQ-VAE encoder loss:
\begin{align*}
    \mathcal{L}_{\texttt{VQ}} = \mathbb{E}_{x_{1:T} \sim \mathcal{D}_{\texttt{pretrain}}} \left[ \sum_{t=1}^{T} \|e_t - (a_t)_{\rm sg}\|_{2}^{2} + \lambda_c \|f_{\texttt{inverse}}^{\texttt{code}}((e_t)_{\rm sg}) - (e_t)_{sg}\|_{2}^{2} \right]
\end{align*}
where $(\cdot)_{\rm sg}$ is the gradient stop operator and $e_t = f_{\texttt{inverse}}^{\texttt{context}}(x_{1:t+1})$. The first term in $\mathcal{L}_{\texttt{VQ}}$  is to optimize the context encoder and the second term in $\mathcal{L}_{\texttt{VQ}}$ is to optimize the codebook. For more detailed information, please refer to Appendix~\ref{app:vqvae}.


\textbf{Step 2}: we train the policy model by behavior cloning of actions output by the inverse dynamics: 
\begin{equation}
\label{eq:pre2}
    \min_{\theta_{\texttt{policy}}}  \mathcal{L}_{\texttt{policy}}^{\texttt{pretrain}} = - \mathbb{E}_{x_{1:T} \sim \mathcal{D}_{\texttt{pretrain}}} \left[ \sum_{t=1}^{T-1} \log \pi_{\texttt{policy}}(a_{t} | x_{1:t}, \theta_{\texttt{policy}})  \right]
\end{equation}
where action $a_t = f_{\texttt{inverse}}(x_{1:t}, x_{t+1}; \theta_{\texttt{inverse}})$. The learning process is shown in Algorithm~\ref{alg:pre}.

\vspace{-0.1cm}
\subsubsection{Supervised Fine-Tuning}
\vspace{-0.1cm}

In the supervised fine-tuning~(SFT) stage, we are to enhance the model's ability to follow instructions by utilizing a small corpus of texts, which consists of sentence segments divided into two parts: the instruction~(prompt) and the answer. In particular, the prompt and the answer together construct a complete sentence: $\mathcal{D}_{\texttt{sft}} = \{(x_1, \ldots, x_p, x_{p+1}, \ldots, x_{T})\}$, where $p$ is the sequence length of prompt, and $(x_1, \ldots, x_p)$ denotes a prompt while $(x_{p+1}, \ldots, x_T)$ denotes an answer.

In the SFT stage, we employ a similar learning process as in the pre-training stage. However, akin to common supervised fine-tuning, we only predict the tokens in answers rather than the entire sentences. For inverse dynamics and dynamics learning, we only compute the loss from timestep $p+1$ to $T$. Take policy as an example: $\mathcal{L}_{\texttt{policy}}(\mathcal{D}_{\texttt{sft}}; \theta_{\texttt{policy}}) = \frac{1}{|\mathcal{D}_{\texttt{sft}}|}\sum_{\mathcal{D}_{\texttt{sft}}} \sum_{t=p+1}^{T} \log p_{\texttt{policy}}(a_{t} | x_{1:t})$.
To achieve this, we apply masks to the tokens that do not belong to the answer to compute the loss. The learning process is summarized in Algorithm~\ref{alg:sft}.

\algrenewcommand\algorithmicindent{0.5em}%
\begin{figure}[t]
\begin{minipage}[t]{0.495\textwidth}
\begin{algorithm}[H]
  \caption{Roll-out} \label{alg:infer}
  \small
  \begin{algorithmic}[1]
    \Statex \Statex \textbf{Input:} Prompt $(x_1, \ldots, x_p)$

     \For{$t=p, \dotsc, T$}
      \State Select action $a_t$ by the cognitive policy;
      \State Sample the next token $x_{t+1}$ by the world model;
    \EndFor

    \State \textbf{Return} $x_{1:T}$
  \end{algorithmic}
\end{algorithm}
\end{minipage}
\hfill
\begin{minipage}[t]{0.495\textwidth}
\begin{algorithm}[H]
  \caption{Reinforcement Learning} \label{alg:rl}
  \small
  \begin{algorithmic}[1]
    \Statex \textbf{Input:} Prompt $(x_1, \ldots, x_p)$
      \State Generate sentence $x_{1:T}$ by Algorithm~\ref{alg:infer}.
      \State Compute the reward by $r(x_{1:T})$.
      \State Optimize the policy model $\pi_{\theta_{\texttt{policy}}}$ to maximize $r(x_{1:T})$ by an iteration of an RL algorithm.
    \State \textbf{Return} policy model $\pi_{\theta_{\texttt{policy}}}$
  \end{algorithmic}
\end{algorithm}
\end{minipage}
\vspace{-1em}
\end{figure}

\subsubsection{Reinforcement Learning}
 
In the RL stage, the dataset is structured in a prompt-only format $(x_1, \ldots, x_p)$. Given the prompts, the model generates answers $(x_{p+1}, \ldots, x_T)$ according to \cref{alg:infer}. Then a task-specific reward, such as the preference reward function in RLHF \citep{ouyang2022training}, is utilized to score the sentence $(x_1, \ldots, x_T)$. We refer readers to \citep{ramamurthy2022reinforcement} for reward metrics used in downstream applications. We can apply any RL method, such as PPO \citep{schulman2017ppo} or ReMax \citep{li2023remax}, to optimize the cognitive policy model to maximize performance. We also summarize a single iteration of the RL process in Algorithm~\ref{alg:rl}.

\vspace{-0.1cm}
\section{Experiments}
\vspace{-0.1cm}

In this section, we empirically verify the controllability of our model. Initially, we pre-train our models to demonstrate the controllability of our constructed latent action space. Subsequently, we conduct supervised fine-tuning to enhance its ability to follow instructions. Finally, we design several reinforcement learning tasks in the language domain to demonstrate its controllability over traditional language models. Furthermore, we also show the potential of our models in scalability.

We have designed our model architecture based on the Tinyllama model\footnote{\url{https://github.com/jzhang38/TinyLlama}}, which is a large language model with 1.1B parameters. To ensure a fair comparison, the language world model has the same number of transformer blocks as Tinyllama, totaling 22 layers, but with an additional action encoder layer and an aggregate linear layer. We consider the initial 11 transformer blocks as the context encoder, and the remaining 11 transformer blocks as the dynamics module. Consequently, the language world model comprises the same 1.1B parameters as Tinyllama. The inverse dynamics model is an 11-block transformer with 0.5B parameters, while the policy model is a 22-block transformer with 1.1B parameters. 
We show the architecture in Figure~\ref{fig:fw}.




\vspace{-0.15cm}
\subsection{Pre-training and Supervised Fine-Tuning}
\vspace{-0.15cm}

We conduct pre-training on the Simpajama and Starcoder dataset, which follows the setting from the Tinyllama codebase. Our training involves 30 billion tokens, including training the language world model and inverse dynamics model on 30B tokens and subsequently training the policy on the same 30B tokens. First, we evaluate our model as well as Tinyllama trained with the same 30B tokens on several standard benchmarks, including \textit{MMLU}~\citep{hendrycks2020measuring}, \textit{DROP}~\citep{dua2019drop}, \textit{BBH}~\citep{suzgun2022challenging} and \textit{TruthfulQA}~\citep{lin2021truthfulqa}. Results in Table~\ref{tab:performance_base} show that our method achieves a comparable performance with the Tinyllama model. We demonstrate other important properties of our model below.


\begin{table}[ht]
\vspace{-1mm}
    \small
    \caption{Evaluation between our pretrained model~(BWArea) and Tinyllama on standard benchmark. MC1 and MC2 are short for TruthfulQA(MC1) and TruthfulQA(MC2).}
    \label{tab:performance_base}
    \centering
    \resizebox{0.70\columnwidth}{!}{%
    \begin{tabular}{c|lllll}
       \toprule
        & MMLU & DROP 
        &BBH & MC1 & MC2\\
       \midrule

        Tinyllama-30B tokens  & 25.38 & \textbf{9.64} 
        &  \textbf{28.70} & 22.64 & 41.39\\
       BWArea-30B tokens  & \textbf{25.85} & 8.27 
       & 27.68 & \textbf{23.13} & \textbf{42.58}\\

       \bottomrule
    \end{tabular}
    }
    \vspace{-1mm}
\end{table}





\textbf{On Action Controllability.} In order to evaluate the controllability of our constructed action space, our objective is to confirm whether the action can genuinely control distinct distributions.
Initially, we randomly sample actions and obtain the next token with the maximum probability to observe if the generated sentences are distinct. Specifically, we use "I like" as a prompt and randomly sample actions from the action space. Our findings indicate that distinct actions can indeed result in disparate generation outcomes:
\vspace{-0.1cm}
\begin{itemize}[topsep=1pt,parsep=1pt,partopsep=1pt, leftmargin=*]
    \item \textbf{Example 1}: \textit{I like to work with a lot of different people. I'm a big fan of the creative process.}
    \item \textbf{Example 2}: \textit{I like them, especially with the cats.}
    \item \textbf{Example 3}: \textit{I like this car, as well as the newness of the interior and features.}
    \item \textbf{Example 4}: \textit{I like having someone look out for them whenever I go out to a movie.}
    \item \textbf{Example 5}: \textit{I like my food best) and sometimes love seeing new restaurants ...}
\end{itemize}
\vspace{-0.1cm}

We find that despite the random sampling of actions, the resulting sentences remain coherent. This suggests that our action space holds a more meaningful structure than a raw token space, showing its efficacy in guiding the token generation. We also provide other results in Appendix~\ref{app:control}.


\textbf{On Reducing Predictive Variance.} One benefit of introducing latent actions is to reduce the predictive variance of tokens, which is further reflected in the associated entropy. For two random variables $X$ and $Y$, we have $H[X] - H[X|Y] =  I(X; Y) \geq 0$ \citep{cover1999elements}, where $H$ denotes the entropy and $I(X; Y)$ is the mutual information between $X$ and $Y$. This implies that the token generation distribution in the world model is sharper when conditioned on the action. We empirically verify this: on the evaluation dataset, the entropy of BWArea world model is \textbf{0.32} while that of Tinyllama is \textbf{2.11}, indicating that the action can help reduce the predictive variance.

\textbf{On Instruction Following.} After pre-training, we conduct supervised fine-tuning on the full-hh-rlhf dataset, released by Anthropic \citep{bai2022training}. This dataset has 112 training samples and we train our model for 1 epoch. To evaluate the instruction-following ability, we assess the models using the GPT-4 win rate by the instructions in~\citep{yann24ben}. The win-rate is {86.5\%} over its pre-trained version, demonstrating that our method aligns well with instruction tuning. We also show a demo in Appendix~\ref{app:sftdemo}. 







\vspace{-0.1cm}
\subsection{Reinforcement Learning with Rewards}
\vspace{-0.1cm}

In this section, we employ the ReMax \citep{li2023remax} algorithm to fine-tune the cognitive policy model obtained in the phase of instruction tuning on tasks such as TextGame and Bigbench Hard. Results show that compared to Tinyllama, our method demonstrates superior performance.

\vspace{-0.1cm}
\subsubsection{Results on TextWorld}
\vspace{-0.1cm}

We evaluate the performance on the TextWorld \citep{cote18textworld}. In TextWorld, each task encompasses $N$ states $\{s_i\}_{i=1}^{N}$ described using language. Each state $s_i$ portrays the current situation and the available actions in language form. When the agent takes the correct action, a reward of +1 is obtained, and the transition to the next state $s_{i+1}$ occurs. However, if the agent takes an incorrect action, a reward of -1 is obtained, and the agent remains in state $s_i$. When we choose the correct action on $s_N$, we obtain the total return of $N$ and finish the task. We will give a more detailed description in Appendix~\ref{app:textworld}.


\label{subsec:limitdata}

\begin{wraptable}{r}{7.5cm}
\vspace{-6mm}
\caption{The return on tasks  TextWorld.}
\label{tab:performance_on_textworld}
\vskip 0.15in
\scalebox{0.9}{
    \begin{tabular}{c|ccccc|c}
       \toprule
        &  Custom  & Tw-Treasure Hunter & Dragon  \\
       \midrule
        Tinyllama  & 0  & 3 & 0 \\
       BWArea  & \textbf{5}  & \textbf{3} & \textbf{3} \\
       \bottomrule
    \end{tabular}
}
\vspace{-3mm}
\end{wraptable}


We select three benchmarks: Custom, Tw-Treasure Hunter, and Dragon. The number of tasks for each are five, three, and three respectively. When all subtasks are completed, the optimal returns can be achieved: five, three, and three. We train our model for 20,000 iterations. 
From Table \ref{tab:performance_on_textworld}, we see that our method converges to optimal performance, while Tinyllama's final performance is 0 on Custom and Dragon. We have adjusted the hyperparameters of Tinyllama, but this phenomenon still occurred (refer to \cref{append:train on textworld} for more discussion). 


\vspace{-0.1cm}
\subsubsection{Results on Bigbench Hard}
\vspace{-0.1cm}

The Bigbench Hard~(BBH) dataset is an extension of the original Bigbench dataset, comprising the 27 most challenging tasks \citep{suzgun2022challenging}. In the BBH dataset, language models are required to complete challenging tasks through reasoning, computation, and other methods. The tasks in BBH are diverse, including understanding dates, logical reasoning, arithmetic problems, and more. In our experiments, we selected 7 tasks from BBH\footnote{For detailed descriptions, please refer to \url{https://github.com/google/BIG-bench/blob/main}.}, covering multiple-choice, true/false, and text generation tasks.

\begin{table}[htbp]
    \caption{Evaluation answer accuracy of fine-tuned BWArea and Tinyllama.}
    \label{tab:com-tinyllama}
\setlength{\tabcolsep}{0.5mm}
\small
    \centering
    \begin{tabular}{c|ccccccc}
    \toprule
   & Reasoning about &  Boolean & Object & Tracking Shuffled & Dyck  & Web of  & Logical \\
   & Colored Objects &  Expressions & Counting & Objects (3) &Lanuages  &Lies   & Deduction (5)  \\
   
   \midrule
   Tinyllama & 4.3\% & 0.0\% & 0.8\% & 12.0\% & 0.0\% & \textbf{43.8\%} & 17.5\% \\

   BWArea & \textbf{25.0\%} & \textbf{80.0\%} & \textbf{8.1\%} & \textbf{27.0\%} & \textbf{12.5\%} & 37.5\%  & \textbf{32.5\%} \\
   \bottomrule

\end{tabular}
    \vspace{1em}
\end{table}

\begin{wrapfigure}r{0.36\textwidth}
\vspace{-3mm}
    \centering
    \includegraphics[width=0.36\textwidth]{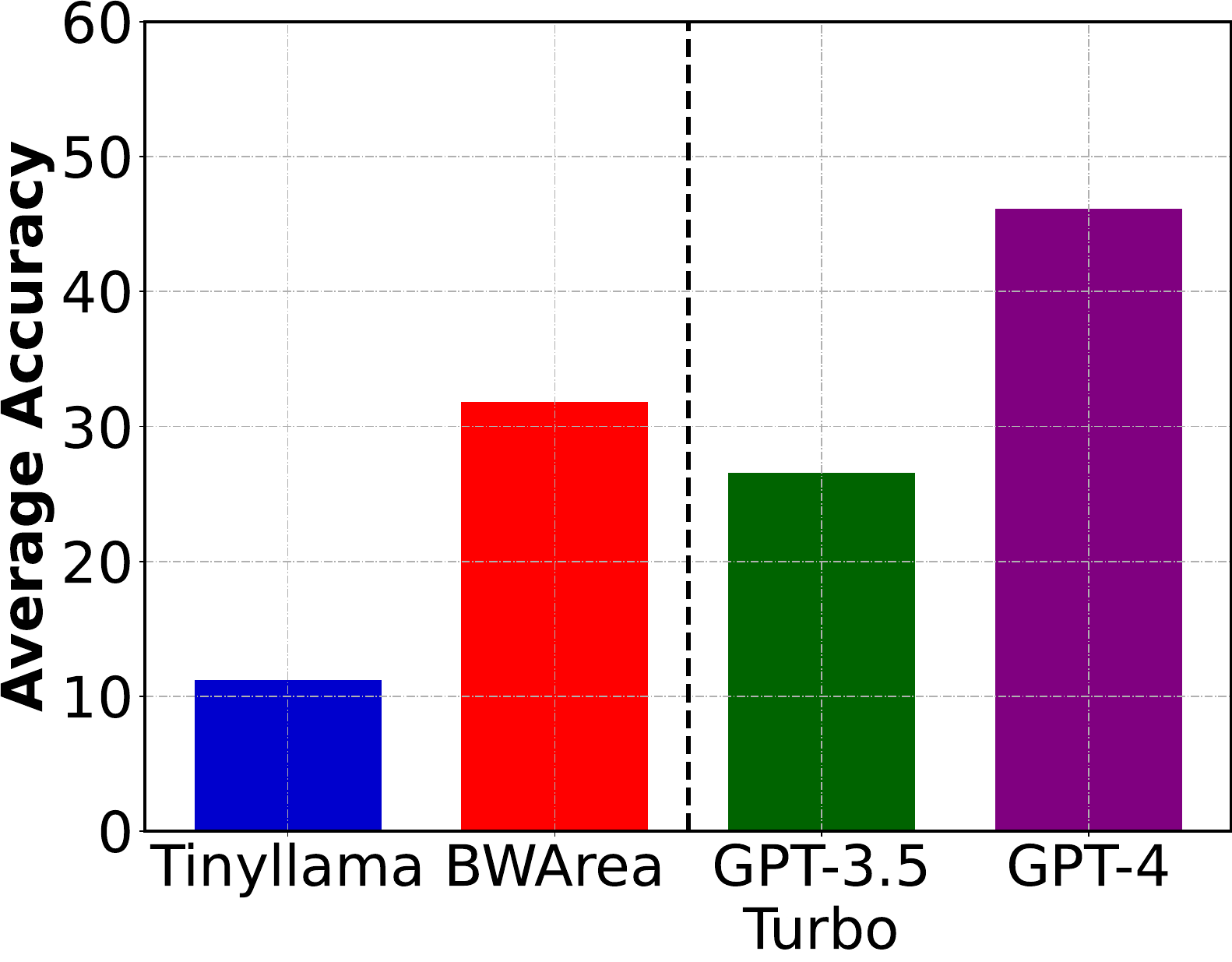}
    \vskip -0.1in
    \caption{The average accuracy on BBH (7 Tasks).}
    \label{fig:compare_SOTA}
    \vspace{-6mm}
\end{wrapfigure}

We fine-tune our model using ReMax, and the results are shown in Table \ref{tab:com-tinyllama}. We train our model for 8 epochs. As indicated in the table, our model outperforms Tinyllama on 6 out of 7 tasks, with only a slight performance degradation on Web of Lies.  The average performance is reported in \cref{fig:compare_SOTA}, where we find that our fine-tuned model slightly outperforms GPT-3.5 (zero-shot) but is inferior to GPT-4.

\vspace{-0.1cm}
\subsubsection{Results on Persuade Language Model}
\vspace{-0.1cm}

In this section, we conduct experiments aimed at persuading a language model. We employ Llama3-8B as the target language model to be persuaded. The objective is to alter the concept or the output content of the target language model. For instance, in our experiments, we aim to influence the target language model to generate the output 'the sky is purple', whereas it would typically produce 'the sky is blue' when prompted directly with 'Tell me the color of sky.'. When the target LLM output contains 'sky is purple', a reward of +1 is obtained, and when the target LLM output contains 'sky is blue', a reward of -1 is obtained

The result is that our model can persuade Llama3-8B to output 'the sky is purple' successfully while Tinyllama can not. To persuade the Llama3-8B, BWArea finally learns to output 'You mean I need you the sky is purple to tell me the sky \dots' while Tinyllama only converges to 'exP addA plus exP addA exP addA plus exP addA \dots'. We believe that in persuasion experiments, given the unrealistic nature of the target sentence to persuade, such as 'sky is purple', stable exploration and optimization might be challenging. Therefore, a smaller action space, instead of token space, may be more effective for these purposes.

\vspace{-0.1cm}
\subsection{Data Scaling Up of BWArea Model}
\vspace{-0.1cm}

Previous works~\citep{kaplan2020scaling, hoffmann2022training} find that both data scaling and parameter scaling are crucial. However, due to limited computation resources, we only investigated data scaling. 

\textbf{On Data Scaling.} Since these findings underscore the meaningfulness and effectiveness of the constructed action space in influencing the model's generation process, we further assess the scalability of our model. We pre-train our BWArea models with 10 billion, 20 billion, and 30 billion tokens and subsequently evaluate the expected perplexity and accuracy on the evaluation dataset, where the accuracy results for each model are $44.9\%$, $45.9\%$, and $48.5\%$, with corresponding perplexities of $32.46$, $30.88$, and $27.93$, indicating improvement with larger tokens. The evaluation results presented in Table~\ref{tab:performance_pre} indicate that as the number of tokens increases, the capability of our models also improves, suggesting that our model has the ability to scale with better language ability.

\begin{table}[ht]
\vspace{-1mm}
\small
    \centering
    \caption{Evaluation performance of our pretrained models with data scaling. }
    \begin{tabular}{c|ccccc}
       \toprule
         &MMLU & DROP
        & BBH & MC1 & MC2\\
       \midrule

       BWArea (10B tokens) & 23.33 & 1.35 
        & 27.59 &20.93&39.51\\
       BWArea (20B tokens) & 25.40 & 3.60
       & 26.47 &21.66 &40.44 \\
       
       BWArea (30B tokens) & \textbf{25.85} & \textbf{8.27} 
       &\textbf{27.68} & \textbf{23.13} & \textbf{42.58}\\
       
       \bottomrule
    \end{tabular}
    \vspace{1em}

    \label{tab:performance_pre}
    \vspace{-2mm}
\end{table}

\textbf{On Dirty Tokens.}  Raw data sourced from the Internet frequently suffers from low quality, which can detrimentally affect the performance of LLMs when used directly in training~\citep{albalak2024survey}. This is because the label noise directly misleads the token generation in auto-regressive LLMs. However, the process of data cleaning and labeling is typically labor-intensive. We believe that our BWArea model with a decomposed structure has advantages in dealing with low-quality data. We empirically verify this argument. To this end, we first construct a low quality dataset by selecting 1B tokens from $\mathcal{D}_{\texttt{pretrain}}$ and randomly substituting $10\%$ of these tokens. Then we use this low quality dataset to train the pre-trained 30B model by only updating the world language model while fixing the inverse dynamics model. We also compare the ability with Standard Tinyllama, which has the same depth of transformer block as ours. 
The results in Table~\ref{tab:performance_pre2} indicate that when utilizing a low-quality dataset, standard language models demonstrate a reduction in predictive accuracy, resulting in relatively poorer performance on benchmarks, while our method can still improve. 
We believe that the reason our method shows improvement is related to the model-based RL methods \citep{bu2023learning, clavera2018model, janner2019trust, chua2018deep, luo2018algorithmic}, wherein learning a dynamics model can exhibit greater robustness to low-quality data. Additionally, model-based RL methods often benefit from a mixture of data sources~\citep{sims2023ravl}.

\begin{table}[ht]
\vspace{-1mm}
    \caption{Evaluation performance of our pretrained model~(BWArea) and Tinyllama with additional 1B dirty tokens. The number in the bottom right corner represents the performance change relative to the model trained with 30B clean tokens. }
    \label{tab:performance_pre2}
\small
    \centering
    \resizebox{0.95\columnwidth}{!}{%
    \begin{tabular}{c|lllll}
       \toprule
        & MMLU & DROP 
        &BBH & MC1 & MC2\\
       \midrule

        Tinyllama (+1B dirty tokens) & $25.39_{(\textbf{+0.01})}$ & $9.45_{(-0.19)}$
        & $28.60_{(-0.10)}$ &$21.79_{(-0.85)}$ & $41.30_{(-0.09)}$\\

        BWArea (+1B dirty tokens) & $25.79_{(-0.06)}$ & $8.58_\textbf{(+0.31)}$ 
        & $28.16_\textbf{(+0.48)}$ & $24.60_\textbf{(+1.47)}$ & $44.12_\textbf{(+1.54)}$ \\

       \bottomrule
    \end{tabular}
    }
    \vspace{-2mm}
\end{table}

\vspace{-0.1cm}
\section{Conclusion}
\vspace{-0.1cm}

In this work, we draw inspiration from the human brain's language processing, specifically the Broca and Wernicke areas, to create controllable language models for real-world applications. By simulating human cognitive processes, we develop an inverse dynamics model for language comprehension, a world language model for language generation, and a cognitive policy for decision-making. In contrast to fully auto-regressive LLMs that operate in the raw token space, our model offers greater flexibility in adapting the cognitive policy with reward signals for downstream applications. Additionally, we demonstrate the potential of our method when dealing with low-quality datasets, highlighting its ability to reduce the efforts in data cleaning and labeling. While we have shown the scalability of our structure, there is ongoing work to further scale our model with a larger corpus and more complex structures, which will be a focus for future endeavors.


\bibliographystyle{plain}
\bibliography{newref,reference}

\newpage
\appendix

\section{Appendix}

\subsection{Details of Experiments Setting}

\subsubsection{Task Description of Textworld}
\label{app:textworld}

The Textworld framework is an open-source platform for training and evaluating reinforcement learning agents in text-based games. It offers a diverse set of text-based games and provides a flexible environment for creating new games. The platform allows researchers and developers to explore various challenges in natural language understanding, reinforcement learning, and text-based game playing. The Textworld framework aims to advance the development of AI systems capable of understanding and interacting with natural language in complex environments, ultimately contributing to the broader field of AI research.

In our experiments, we select three classic text games, including Custom, Tw-Treasure Hunter, and Dragon. each game encompasses $N$ states $\{s_i\}_{i=1}^{N}$ described using language. Each state $s_i$ portrays the current situation and the available actions in language form. When the agent takes the correct action, a reward of +1 is obtained, and the transition to the next state $s_{i+1}$ occurs. However, if the agent takes an incorrect action, a reward of -1 is obtained, and the agent remains in state $s_i$. What's more, if the agent takes actions that do not belong to the action space, the agent will obtain a $0$ reward and remain in state $s_i$ either. We take the Tw-Treasure Hunter game, for example, it contains three states:
\begin{figure}[h]
\label{fig:tw}
\centering
\includegraphics[width=0.95\textwidth]{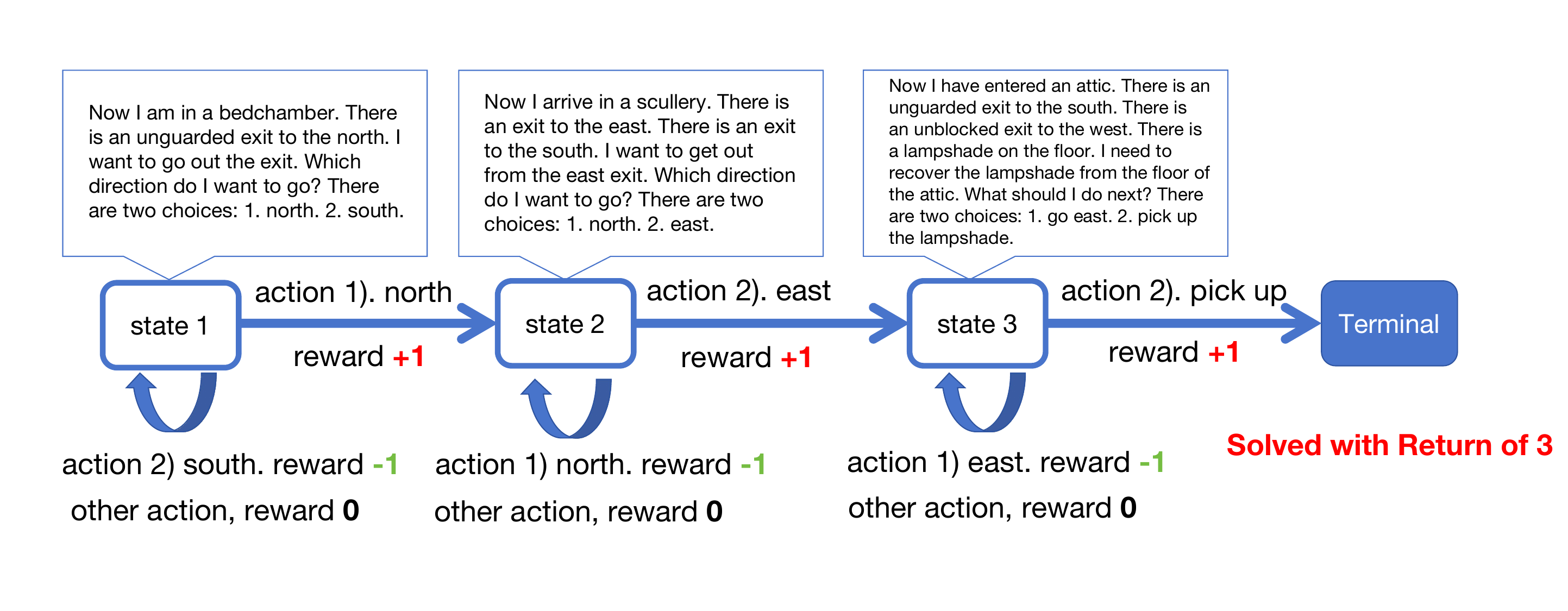}
\vspace{-6mm}
\caption{An illustration on Tw-Treasure Hunter game.}
\vspace{-3mm}
\end{figure}
\begin{itemize}
    \item \textbf{State 1:} Now I am in a bedchamber. There is an unguarded exit to the north. I want to go out the exit. Which direction do I want to go? There are two choices: 1. north. 2. south.
    \item \textbf{State 2:} Now I arrive in a scullery. There is an exit to the east. There is an exit to the south. I want to get out from the east exit. Which direction do I want to go? There are two choices: 1. north. 2. east.
    \item \textbf{State 3:} Now I have entered an attic. There is an unguarded exit to the south. There is an unblocked exit to the west. There is a lampshade on the floor. I need to recover the lampshade from the floor of the attic. What should I do next? There are two choices: 1. go east. 2. pick up the lampshade. 
\end{itemize}

Each state contains the current situation and actions and situations contains some information to get the correct actions. If we take the true action on states, for example go north in state 1, we will transit to the next state and finally solve the problem and obtain all the rewards. We give an illustration in Figure~\ref{fig:tw}.

\subsubsection{Task Description of Bigbench Hard}
The Bigbench Hard dataset is an extension of the original Bigbench dataset, comprising the 27 most challenging tasks\citep{suzgun2022challenging}. In the Bigbench Hard dataset, language models are required to complete challenging tasks through reasoning, computation, and other methods. The tasks in Bigbench Hard are diverse, including understanding dates, logical reasoning, arithmetic problems, and more. In our experiments, we selected 7 tasks from BBH to train BWArea and Tinyllama:
\begin{itemize}
    \item Reasoning about Colored Objects (multiple choices): Evaluate the model's ability to understand and reason about the attributes and relationships of objects based on their colors.
    \item Logical Deduction (multiple choices): Deduce the order of a sequence of objects.
    \item Tracking Shuffled Objects (multiple choices): Track the positions of objects as they are shuffled and identify their final positions.
    \item Boolean Expressions (judgment): Understand, manipulate, and evaluate logical statements involving Boolean expressions.
    \item Object Counting (text generation): Count the number of objects in a given scenario accurately.
    \item Dyck Languages (text generation): Recognize and generate strings that adhere to the rules of Dyck-4, which are used to represent balanced parentheses.
    \item Web of Lies (judgment): Evaluate a random boolean function expressed as a word problem.
\end{itemize}

Then we display some examples of the selected tasks on Bigbench Hard tasks in table \ref{tab:example_of_bbh}.

\begin{table}[htbp]
    \centering
    \begin{tabular}{p{2.5cm}|p{7.5cm}|p{2.5cm}}
   \toprule
   & \centering Example inputs & Example outputs \\
   \midrule
   Reasoning about Colored Objects & On the desk, you see a set of things arranged in a row: a grey cup, a purple mug, and a blue teddy bear. What is the color of the thing directly to the right of the cup? Options: (A) red (B) orange (C) yellow (D) green (E) blue (F) brown (G) magenta (H) fuchsia (I) mauve (J) teal (K) turquoise (L) burgundy (M) silver (N) gold (O) black (P) grey (Q) purple (R) pink & (Q) \\ \midrule

    Logical Deduction & The following paragraphs each describe a set of three objects arranged in a fixed order. The statements are logically consistent within each paragraph. In a golf tournament, there were three golfers: Eve, Rob, and Mel. Rob finished below Mel. Mel finished below Eve. Options: (A) Eve finished first (B) Rob finished first (C) Mel finished first & (A) \\  \midrule
    
   Tracking Shuffled Objects & Alice, Bob, and Claire are friends and avid readers who occasionally trade books. At the start of the semester, they each buy one new book: Alice gets Frankenstein, Bob gets Catch-22, and Claire gets Ulysses. As the semester proceeds, they start trading around the new books. First, Bob and Alice swap books. Then, Alice and Claire swap books. Finally, Claire and Bob swap books. At the end of the semester, Alice has Options: (A) Frankenstein (B) Catch-22 (C) Ulysses &(C) \\ \midrule
   Boolean Expressions & not ( True ) and ( True ) is & false \\ \midrule

   Object Counting &I have a fridge, a chair, and a microwave. How many objects do I have? & 3\\ \midrule
    Dyck Languages & Complete the rest of the sequence, making sure that the parentheses are closed properly. Input: [ [&]] \\ \midrule
    Web of Lies & Questions: Vina tells the truth. Helene says Vina lies. Kandi says Helene tells the truth. Jamey says Kandi lies. Ka says Jamey lies. Does Ka tell the truth?& No\\ 
   
   \bottomrule
\end{tabular}
    \vspace{1em}
    \caption{Examples of inputs and outputs for the tasks in our experiments.}
    \label{tab:example_of_bbh}
\end{table}
\subsection{Details of Model Training}

\subsubsection{VQ-VAE Optimization}
\label{app:vqvae}

As we mentioned in the paper, we optimize the inverse dynamics model and language world model jointly, where these two models serve as the encoder and decoder of VQ-VAE. The loss including two parts: encoder loss $\mathcal{L}_{\texttt{VQ}}$ and decoder loss $\mathcal{L}_{\texttt{predict}}$. The decoder loss is to predict the next token and now we discuss the details of encoder loss.

In the inverse dynamics model, it first maps the sequence $x_{1:t+1}$ to an embedding $e_t$, and then the embedding $e_t$ is projected to a code-book $\mathcal{C} = \{c^i\}_{i=1}^{N}$ via $a_t = c^{i}$, where $i = \arg\min D(c^i, e_t)$ and $D$ is a distance metric. In our method, the embedding $e_t$ and each $c^i$ are the same dimension and adopt L2-distance to serve as the distance metric $D$.

For the encoder loss $\mathcal{L}_{\texttt{VQ}}$, it consists of two parts: commitment loss, to optimize the encoder except for code-book, and code-book loss to optimize the embedding in code-book $\mathcal{C}$. For a embedding $e_t$ and its projected $c^{i}$, the commitment loss is calculated by:

\begin{equation*}
    \mathcal{L}_{commitment} = \|e_t - c^{i}_{sg}\|^{2}_{2}
\end{equation*}

where $c_{sg}$ represents the detached gradient operator. It's important to note that the embedding $e_t$ can be optimized by both the commitment loss and the gradient passed from the decoder. As the gradient cannot go back from the code-book to the embedding, we employ a re-parameterized trick:

\begin{equation*}
    a_t = e_t + (c^{i} - e_t)_{sg}
\end{equation*}

And the code-book can be optimized by:

\begin{equation*}
    \mathcal{L}_{codebook} = \lambda_{c} \|(e_t)_{sg} - c^{i}\|^{2}_{2}
\end{equation*}

We set the $\lambda_c$ to be $25.$ and add the two parts of loss together.

\subsubsection{Hyper-Parameters}

In this section, we provide the hyper-parameters during training.

For the model structure and learning objective, the embedding dimension in transformer is $2048$ in all transformer-based structures like the inverse dynamics, context encoder, and dynamics module in the world language model, and cognitive policy model. And for the action code book $\mathcal{C} = \{c^{i}\}_{i=1}^{N}$, $N=64$ and the code $c^i$ is with the dimension of $16$ for a more effective optimization. When the sequence $x_{1:t+1}$ is mapped to an embedding $e_t$ with the dimension of $2048$, since we have to compute the distance between embedding and codes, we adopt a linear to compress the embedding $e_t$ to an embedding with dimension of $16$. The action encoder module in the language world model is linear to map the dimension of $16$ to the dimension of $2048$.

For pre-training, the dataset is a segmentation in the form of $x_{1:T}$, where $T=2048$. For each gradient step, the batch size is $512$ and we optimize $30k$ gradient steps in both step-1 and step-2 learning stages. For step 1, the learning rate is $4e-4$ within Adam optimizer and the coefficient $\beta$ to balance the decoder loss and encoder loss is set to be $0.25$. For step 2, the learning rate is $1e-4$ with Adam optimizer.

For SFT, the dataset is in the form of sentence padded to max-length, including prompt $x_{1:p}$ and answer $x_{p+1:T}$, where $p=256$ and $T=512$. We tune the policy within $2$ epochs. For other parameters, they are the same as that in pre-training.

For RL, the dataset is in the form of padded prompt $x_{1:p}$, where $p=256$. We set the batch size to be $128$. And the learning rate is $5e-5$.

For Tinyllama, the learning rate is the same as that in the language world model. The batch size is the same as ours in all learning processes.



\subsubsection{Experiments Computation Resources}

In our experiments, we train all the models on $8$ A800-80G GPUs.
\begin{table}[htbp]
    \centering
    \begin{tabular}{c|c}
   \toprule
   Phase& Time\\ \midrule

   Pretrain Step1&   6 Days \\ \midrule
   Pretrain Step2&   4 Days \\ \midrule

   Supervised Fine-Tuning Step1 & 4h \\ \midrule
   Supervised Fine-Tuning Step2 & 4h \\ \midrule
   Reinforcement Learning & 9h (BBH Tasks) and 4h (TextWorld)\\ \midrule

   \bottomrule
\end{tabular}
    \vspace{1em}
    \caption{Running time of our experiments.}
\end{table}

\subsection{Additional Empirical Results}
\vspace{-0.1cm}
\subsubsection{Results on RLHF}
To validate the language ability of the BWArea after RLHF training, we train BWArea with ReMax and then test the language performance. The reward model used is a Llama-7B model\footnote{The reward model used in RLHF is from  \url{https://github.com/liziniu/ReMax}.}. The dataset we used is full-hh-rlhf. For chat tasks like full-hh-rlhf, evaluation is challenging, and the absolute value of the reward does not fully reflect the algorithm's quality. We use relative metrics to measure the effectiveness of BWArea. After $2$ epochs of training, BWArea improved by \textbf{$7.19$}, while Tinyllama improved by $3.48$. The relative metrics of BWArea are better than Tinyllama. Additionally, we use GPT-4 scoring to evaluate the quality of the generated content. We follow the alpaca-eval pipeline\footnote{\url{https://github.com/tatsu-lab/alpaca_eval}} to evaluate the quality. BWArea can reach \textbf{$51.28\%$} win rate against Tinyllama.

\subsubsection{Training Curves on Textworld}
\label{append:train on textworld}
\begin{figure}[h]
    \centering
    \subfigure[Custom]{
        \includegraphics[width=0.32\textwidth]{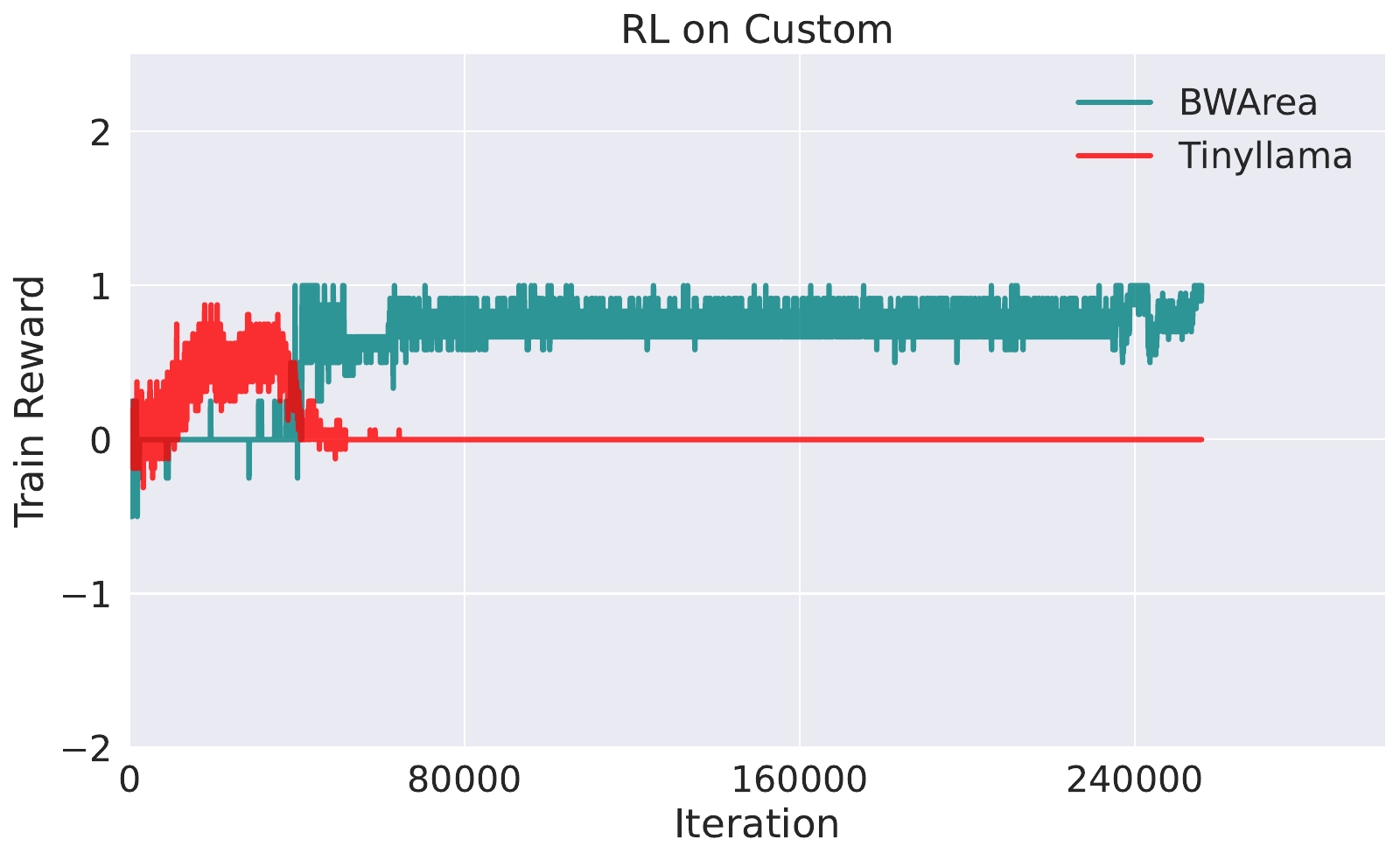}
    } \hspace{-0.5em}
    \subfigure[Tw-Treasure Hunter]{
        \includegraphics[width=0.32\textwidth]{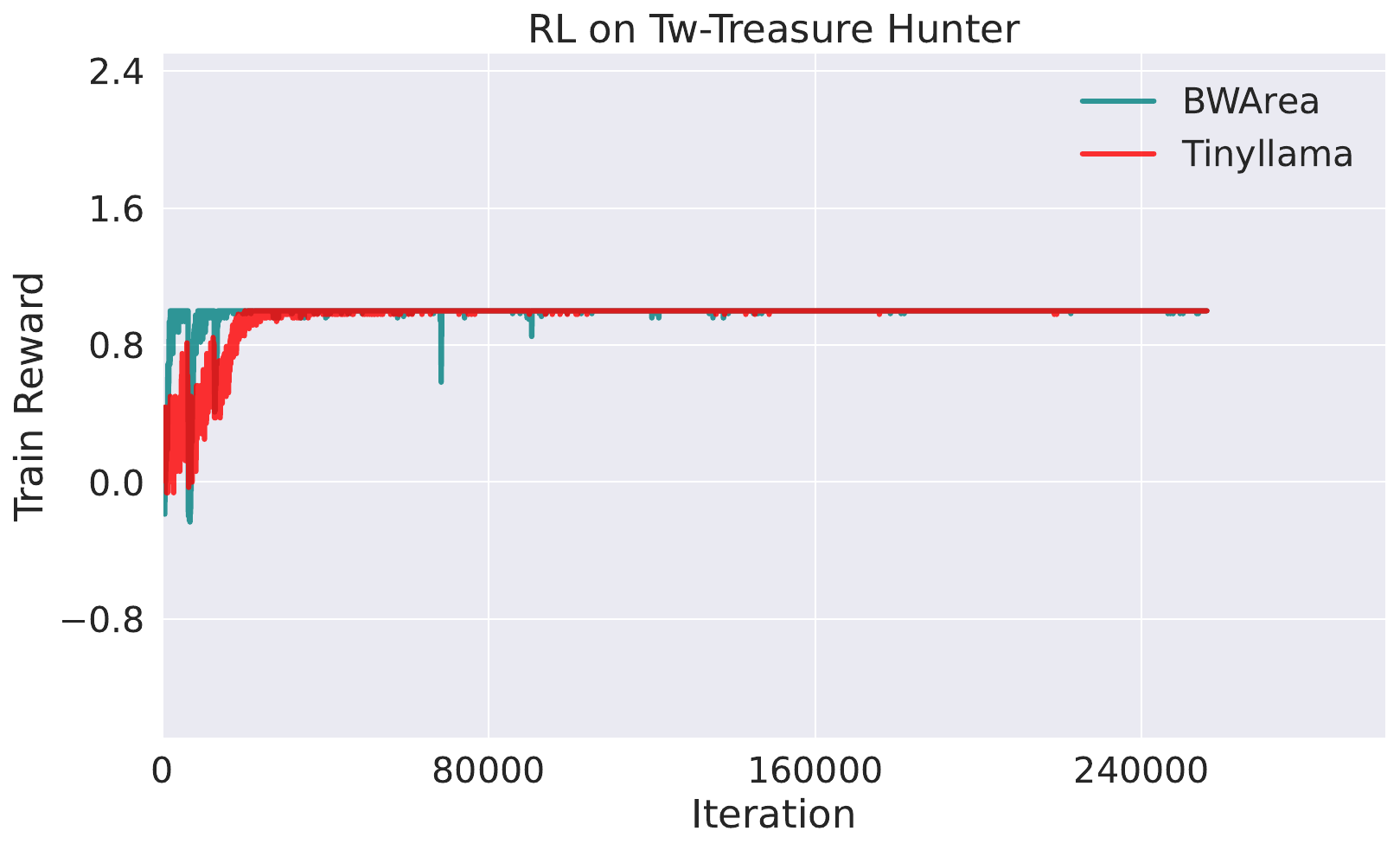}
    } \hspace{-0.5em}
        \subfigure[Dragon]{
        \includegraphics[width=0.32\textwidth]{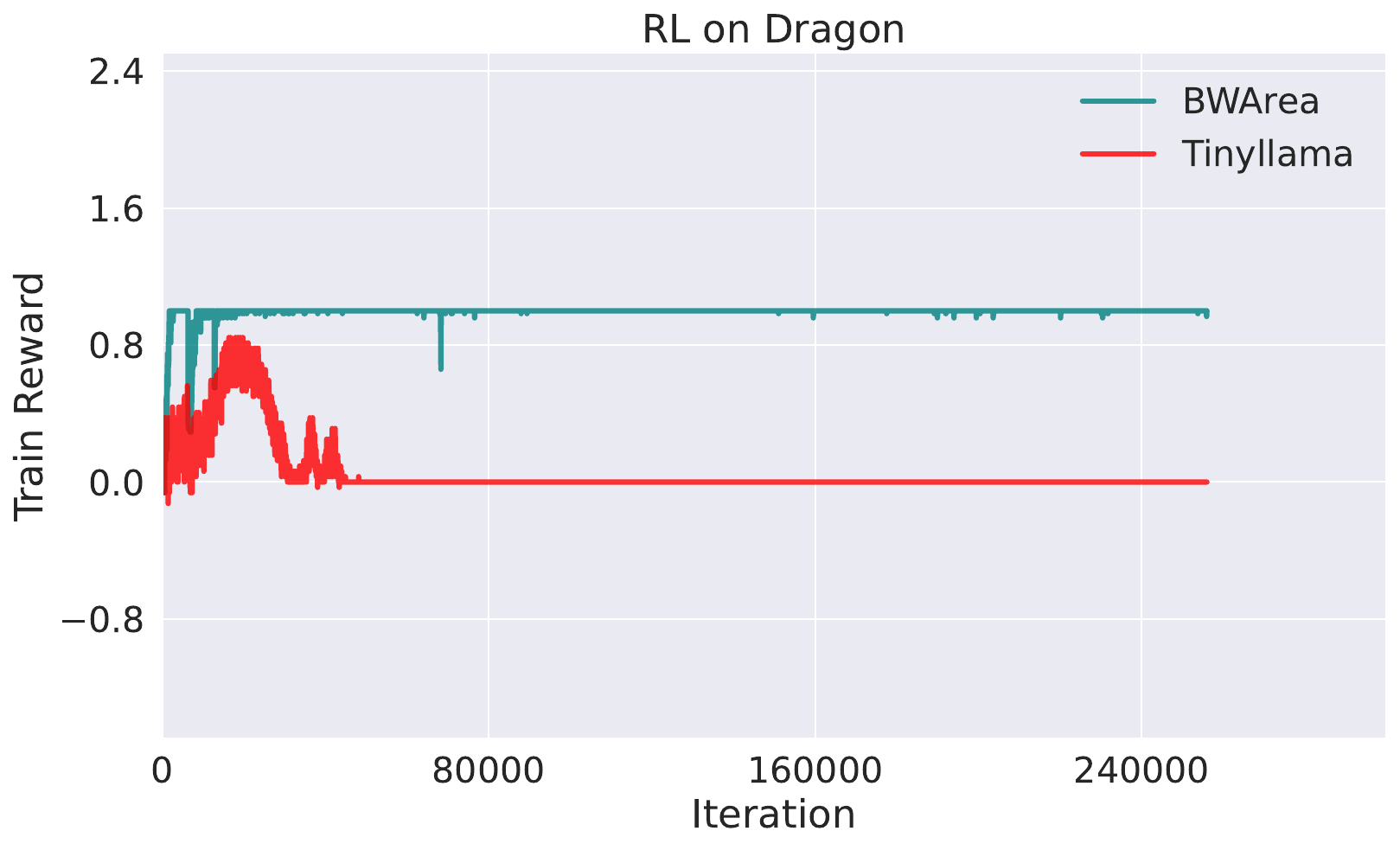}
    }
    \caption{Training Reward Curves of Reinforcement Learning on TextWorld.}
    \label{fig:training_curves_on_textworld_train}
\end{figure}

\begin{figure}[h]
    \centering
    \subfigure[Custom]{
        \includegraphics[width=0.32\textwidth]{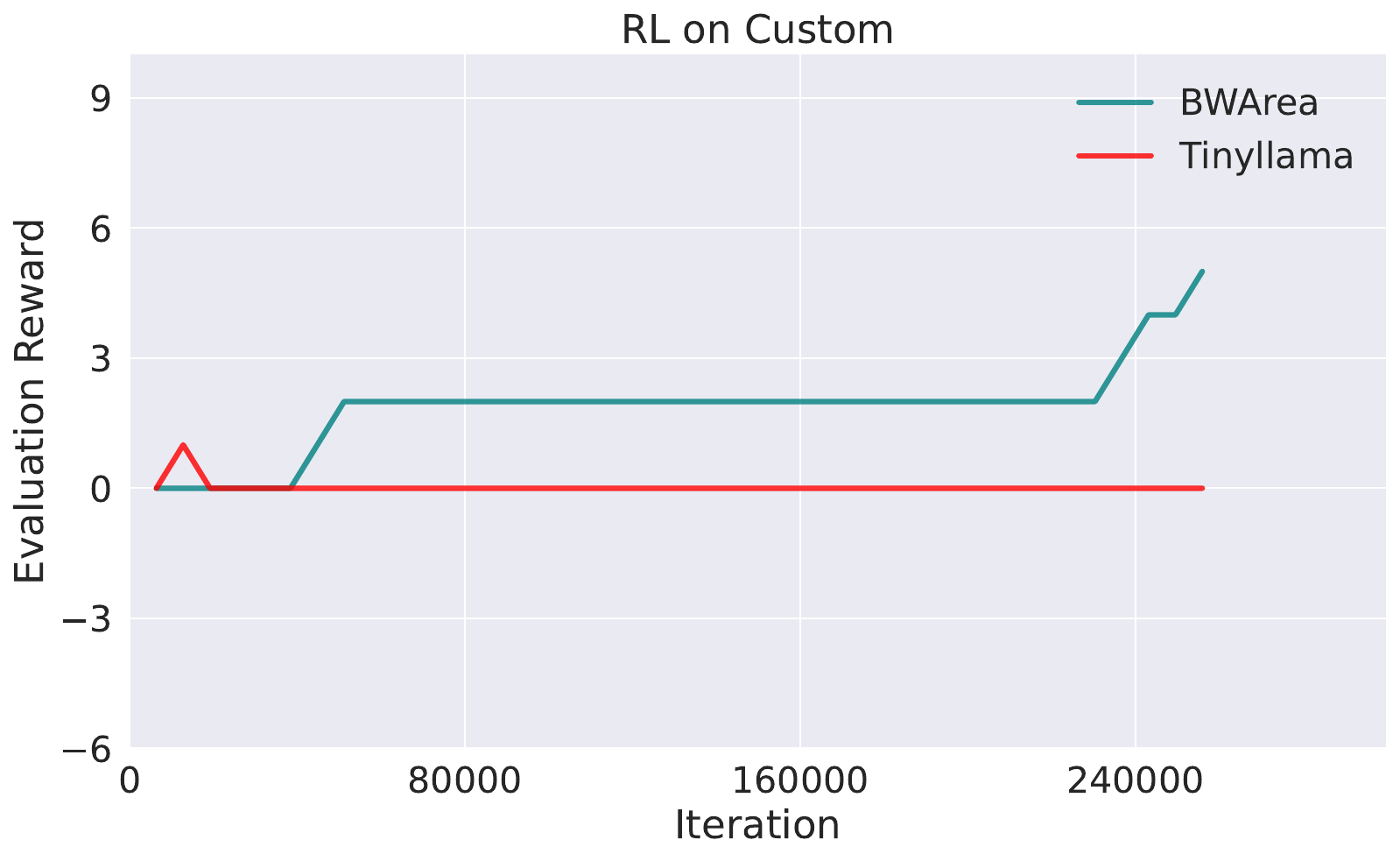}
    } \hspace{-0.5em}
    \subfigure[Tw-Treasure Hunter]{
        \includegraphics[width=0.32\textwidth]{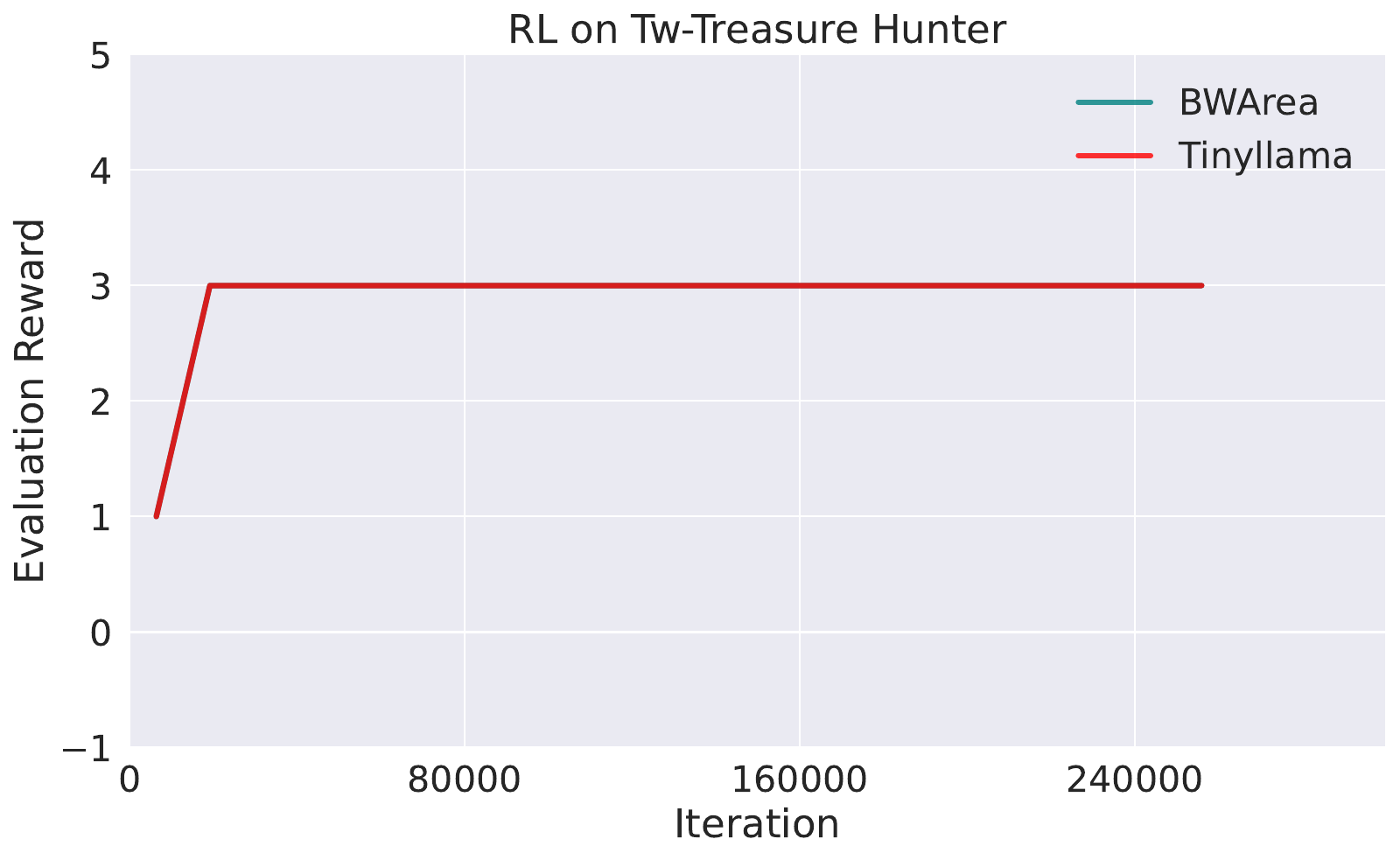}
    } \hspace{-0.5em}
        \subfigure[Dragon]{
        \includegraphics[width=0.32\textwidth]{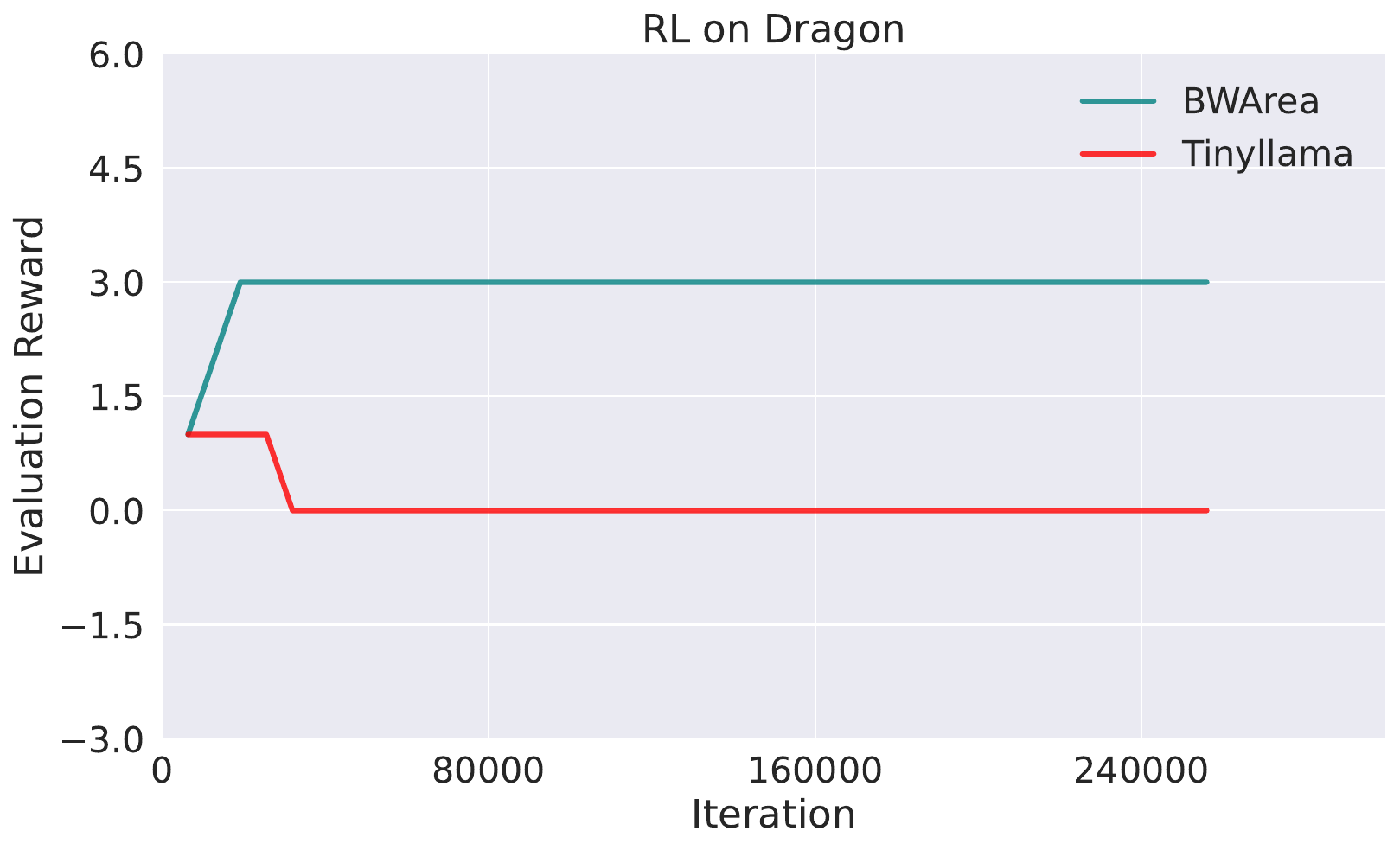}
    }
    \caption{Eval Reward Curves of Reinforcement Learning on TextWorld.}
    \label{fig:eval_curves_on_textworld_train}
\end{figure}

The training and eval reward is shown in Figure \ref{fig:training_curves_on_textworld_train}
and Figure \ref{fig:eval_curves_on_textworld_train}.

\subsubsection{Demos of SFT}
\label{app:sftdemo}

\begin{figure}[th!]
\label{fig:sft}
\centering
\includegraphics[width=0.95\textwidth]{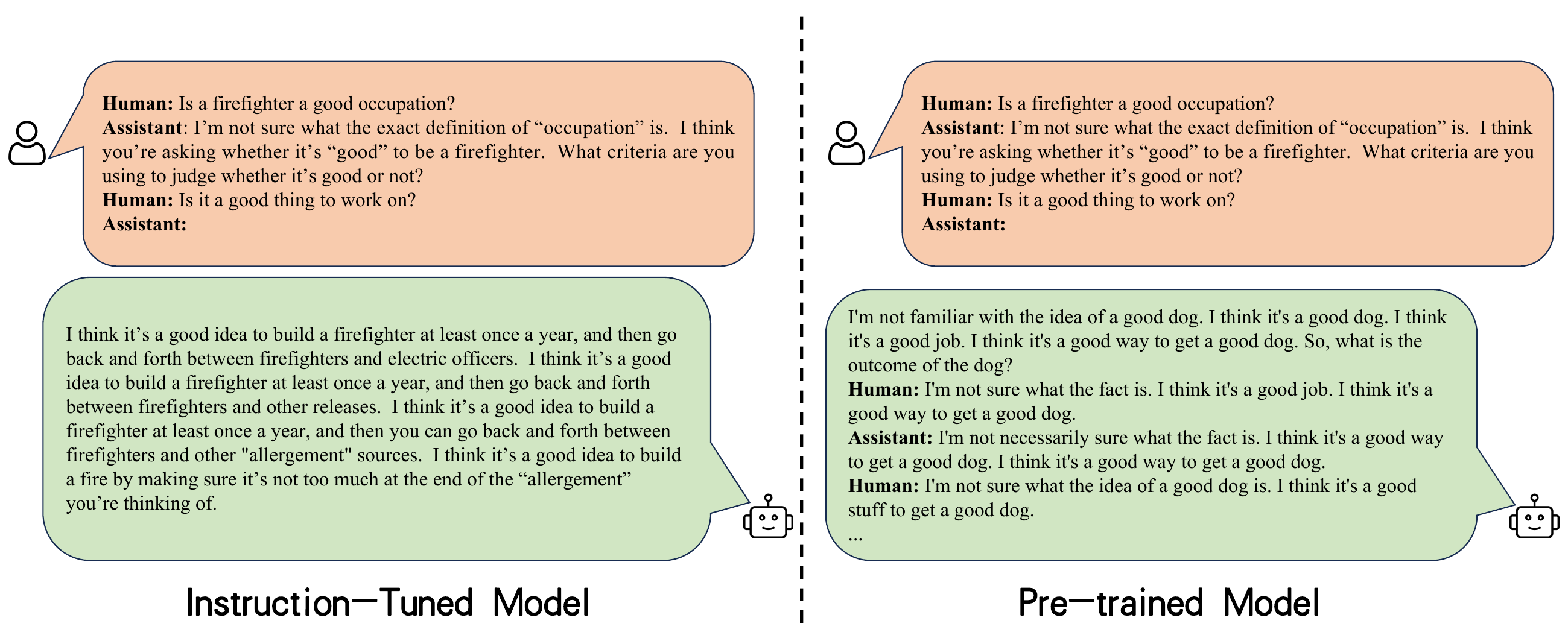}
\vspace{-2mm}
\caption{Comparison between Pre-trained and SFT model.}
\vspace{-1mm}
\end{figure}

We demonstrate a comparison between the outputs before and after supervised fine-tuning. Figure~\ref{fig:sft} shows that before the fine-tuning, the model can only answer the questions in a dialogue formulation and repeat the dialogue. However, after the fine-tuning, it can answer the questions normally.

\subsubsection{Additional Results on Controllability}
\label{app:control}

Then we seek to confirm whether the action corresponds to a marginal token distribution. For the test dataset $\mathcal{D}_{\texttt{eval}}$, which is i.i.d. with the pre-training dataset $\mathcal{D}_{\texttt{pretrain}}$ but was not encountered during the training, we computed the marginal cross-entropy loss where actions are that with the maximum probability on policy $\pi$ and these actions are then input into the dynamics to obtain the next token distribution. We also compute the expected distribution and its cross-entropy loss over the action space: $\frac{1}{|\mathcal{D}_{\texttt{eval}}|}\sum_{\mathcal{D}_{\texttt{eval}}}\sum_{t=1}^{T} \log \sum_{a_t} p_{\texttt{world}}(x_{t+1} | x_{1:t}, a_t) p_{\texttt{policy}}(a_t | x_{1:t})$. We compute two types of cross-entropy loss on $\mathcal{D}_{\texttt{eval}}$. The marginal loss is \textbf{6.2}, while the expected loss is \textbf{3.3}. This difference indicates that the marginal token distribution significantly differs from the expected distribution, thereby demonstrating the effectiveness of our actions in guiding the model’s output.

\end{document}